\pdfoutput=1

\documentclass[11pt]{article}

\usepackage{EMNLP2023}

\usepackage{times}
\usepackage{latexsym}
\usepackage{amsmath}
\usepackage{graphicx}
\usepackage{subcaption}
\usepackage{amsfonts} 
\usepackage{bm} 
\usepackage[frozencache,cachedir=.]{minted}
\usepackage[many]{tcolorbox}
\usepackage{booktabs}

\usepackage[T1]{fontenc}

\usepackage[utf8]{inputenc}

\usepackage{microtype}

\usepackage{inconsolata}

\title{Position Engineering: Boosting Large Language Models through Positional Information Manipulation}

\author{
  Zhiyuan He, Huiqiang Jiang, Zilong Wang, Yuqing Yang, Luna Qiu, Lili Qiu \\
  Microsoft Research \\
  \tt \{zhiyuhe,hjiang,wangzilong,yuqyang,lunaqiu,liliqiu\}@microsoft.com
}

\begin{document}
\maketitle
\begin{abstract}

The performance of large language models (LLMs) is significantly influenced by the quality of the prompts provided. In response, researchers have developed enormous prompt engineering strategies aimed at modifying the prompt text to enhance task performance. In this paper, we introduce a novel technique termed \textit{position engineering}, which offers a more efficient way to guide large language models. Unlike prompt engineering, which requires substantial effort to modify the text provided to LLMs, position engineering merely involves altering the positional information in the prompt without modifying the text itself. We have evaluated position engineering in two widely-used LLM scenarios: retrieval-augmented generation (RAG) and in-context learning (ICL). Our findings show that position engineering substantially improves upon the baseline in both cases. Position engineering thus represents a promising new strategy for exploiting the capabilities of large language models.

\end{abstract}

\section{Introduction}

Recent advancements in Large Language Models (LLMs) have demonstrated significant strides towards achieving artificial general intelligence. These models exhibit a wide range of capabilities, such as in-context learning \cite{brown2020language}, answering questions based on documents \cite{lewis2020retrieval,guu2020retrieval}, solving complex mathematical problems \cite{frieder2024mathematical}, and generating code \cite{romera2024mathematical,ma2023eureka}.

When utilizing LLMs, user prompts are inputted, converted into sequences of tokens, and then processed through multiple attention layers \cite{vaswani2017attention}. These attention layers employ two types of information derived from the token sequences: (i) Semantic information, where the tokens are converted into text embeddings, and (ii) Positional information, where the indices of the tokens are converted into positional embeddings \cite{vaswani2017attention,su2024roformer}. The attention mechanism then combines the semantic and positional information to predict the distribution of the next token in the sequence.

\begin{figure}[htbp]

\begin{subfigure}{\linewidth}
    \centering
    \includegraphics[width=\linewidth]{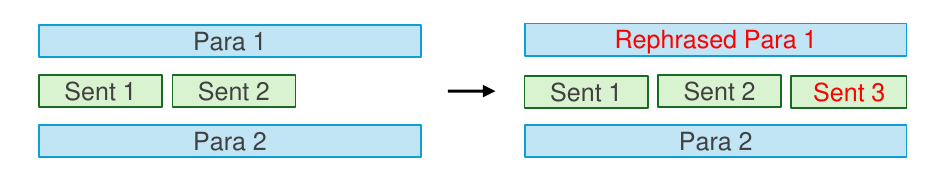}
    \caption{Prompt engineering}
    \label{fig:prompt-engineering}
\end{subfigure}

\begin{subfigure}{\linewidth}
    \centering
    \includegraphics[width=\linewidth]{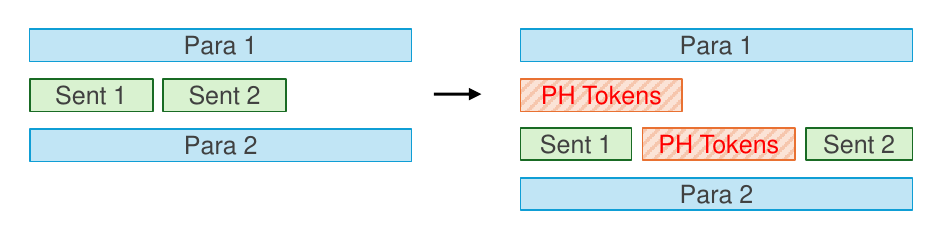}
    \caption{Position engineering}
    \label{fig:position engineering}
\end{subfigure}
\caption{Comparison of prompt engineering and position engineering. "Para" refers to paragraphs, and "Sent" to sentences in prompts. Prompt engineering involves either adding, replacing, or removing paragraphs and sentences from prompts. In contrast, the proposed position engineering maintains the original prompt text but incorporates placeholder tokens instead. These placeholders are not involved in the computation of attention scores, thus the computation overhead is not increased. However, they do hold position indices, thereby affecting the position information of other tokens in the text.}
\label{fig:comparison-prompt-position}
\end{figure}

Extensive research has been conducted on modifying prompt text to alter semantic information, aiming to boost task performances. For instance, few-shot prompting is introduced, enabling LLMs to learn new tasks in an in-context manner \cite{brown2020language}. Moreover, the Chain-of-Thought methodology has been introduced to enhance LLMs' reasoning abilities by prompting them to produce intermediate tokens \cite{wei2022chain,kojima2022large}. Additionally, Automatic Prompt Engineer has been developed to autonomously design the prompting text for better task-specific performance \cite{zhou2022large}.

In this study, we investigate the potential of improving performance by solely modifying positional information, without any semantic information change. For the first time, we reveal that downstream task performance can be significantly enhanced by simply adjusting the positional indices of tokens, without modifying the text itself.

As illustrated in Figure \ref{fig:comparison-prompt-position}, our approach involves the introduction of placeholder tokens to modify positional information. These placeholder tokens do not contribute to the computation of attention scores; however, they do occupy token indices. Consequently, the relative position of other tokens is altered, which could optimize the attention weights among different segments within the prompts. We refer to this approach as \textit{position engineering}, highlighting the exclusive focus on manipulating positional information.

We propose a simple yet effective method based on brutal force to discover the optimal placeholder token number for each downstream task, and experiment it within two prevalent scenarios of LLMs: Retrieval-Augmented Generation (RAG) and In-Context Learning (ICL). Our method significantly enhances performance in both tasks, achieving up to a 15.4\% absolute increase in accuracy for RAG and a 3.6\% absolute increase for ICL. We also discover that the same placeholder number can consistently improves the RAG's performance for different datasets and models. 



In all, our contributions can be summarized as follows:

\begin{itemize}
    \item For the first time, we discover that different downstream tasks' performances can be improved by merely changing the positional information in prompts.
    \item We propose a method to help find a better positional information setting.
    \item We demonstrate that RAG performance can be consistently improved by a universal positional information setting on different datasets and models.
\end{itemize}

\section{Methodology}
\label{sec:Methodology}

\subsection{Preliminary}

In this section, we provide a brief overview of how large language models (LLMs) integrate position information. Let $\{t_i\}_{i=1}^{N}$ represent the input tokens to language models, and let $\{\mathbf{e}_i\}_{i=1}^{N}$ denote the corresponding token embeddings. Initially, the attention layer computes $\mathbf{q}, \mathbf{k}, \mathbf{v}$:
\begin{equation}
\label{eq:general-attention-position}
\begin{split}
    \mathbf{q}_m &= f_q(\mathbf{e}_m, m) \\
    \mathbf{k}_n &= f_k(\mathbf{e}_n, n) \\
    \mathbf{v}_n &= f_v(\mathbf{e}_n, n)
\end{split}
\end{equation}
where $m$ and $n$ are the position indices of tokens. The self-attention is then calculated as follows:
\begin{equation}
\begin{split}
\label{eq:attention_output}
a_{m,n} &= \frac{e^{\frac{\mathbf{q}_m^\mathsf{T}\mathbf{k}_n}{\sqrt{d}}}}{\sum^{N}_{j=1}{e^{\frac{\mathbf{q}_m^\mathsf{T}\mathbf{k}_j}{\sqrt{d}}}}} \\
\textbf{o}_{m} &= \sum_{n=1}^{N}{a_{m,n}\textbf{v}_n}
\end{split}
\end{equation}
where $a_{m,n}$ is a scalar capturing the attention score between $m$-th token in the query and $n$-th token in the value and key sets. $d$ denotes the dimension of the attention layer, and $o_m$ indicates the output for the $m$-th query token.

Absolute positioning is initially introduced by incorporating a positional embedding vector $\mathbf{p}_n$, which is related to $m$ and $n$ \cite{vaswani2017attention}:
\begin{equation}
\begin{split}
    f_q(\mathbf{e}_m, m) = \textbf{W}_q(\mathbf{e}_m + \mathbf{p}_m)\\
    f_k(\mathbf{e}_n, n) = \textbf{W}_k(\mathbf{e}_n + \mathbf{p}_n) \\
    f_v(\mathbf{e}_n, n) = \textbf{W}_v(\mathbf{e}_n + \mathbf{p}_n)
\end{split}
\end{equation}
The $2i$ and $2i+1$ dimension of the positional embedding $\mathbf{p}_n$ is calculated as follows:
\begin{equation}
\begin{split}
    \mathbf{p}_{n,2i} &= sin(n / {10000^{\frac{2i}{d}}}) \\
    \mathbf{p}_{n,2i+1} &= cos(n / {10000^{\frac{2i}{d}}})
\end{split}
\end{equation}

Recently, RoPE adopts the relative position information instead of the absolute information \cite{su2024roformer}. It utilizes a specifically designed matrix $\textbf{R}^d_{i}$, of dimensions $d \times d$ and parameterized by $i$, to modify the query and key vectors in the following manner:
\begin{equation}
\begin{split}
    f_q(\mathbf{e}_m, m) = \textbf{R}^d_{m}\textbf{W}_q\mathbf{e}_m \\
    f_k(\mathbf{e}_n, n) = \textbf{R}^d_{n}\textbf{W}_k\mathbf{e}_m \\
    f_v(\mathbf{e}_n, n) = \textbf{W}_v\mathbf{e}_n
\end{split}
\end{equation}
The matrix $\textbf{R}^d_{i}$ has a unique property, namely $(\textbf{R}^d_i)^\mathsf{T}\textbf{R}^d_j=\textbf{R}^d_{j-i}$, which leads to:
\begin{equation}
\begin{split}
\mathbf{q}_m^\mathsf{T}\mathbf{k}_n = \textbf{e}_m\textbf{W}_q\textbf{R}^d_{n - m}\textbf{W}_k\textbf{e}_n
\end{split}
\end{equation}
Consequently, in Equation (\ref{eq:attention_output}), the model solely focuses on the relative position $n-m$, instead of the absolute positions $n$ and $m$. RoPE has been adopted by recent LLMs, including Llama, Llama2 and Mistral \cite{touvron2023llama,touvron2023llama2,jiang2023mistral}.

\subsection{Altering Position Information in Prompts}
\label{sec:alter-position-information}

The performance of LLMs is significantly influenced by the quality of the prompts used. To enhance the effectiveness of these prompts, researchers have developed a wide range of prompt engineering strategies. This refinement process involves transforming the initial input tokens $\{t_i\}_{i=1}^{N}$ into revised inputs $\{\widehat{t_j}\}_{j=1}^{\widehat{N}}$, which necessitates modifications to the text. For instance, the Zero-shot chain-of-thought technique enhances the reasoning abilities of LLMs by appending the sentence "Let's think step by step." to the prompts \cite{kojima2022large}. 


In this paper, we propose a novel methodology termed "position engineering" to further exploit the capabilities of LLMs. Unlike prompt engineering, position engineering requires no modification to the input tokens themselves.  Instead, it solely modifies the position information utilized in Equation (\ref{eq:general-attention-position}).  Through empirical experiments, we have discovered that such adjustments to position information can significantly improve performance. Formally, we aim at discovering a position editing function, $\tau(\cdot): \mathbb{N} \rightarrow \mathbb{N}$, that boosts LLM performance. This function changes the token position information, which is incorporated into the model as shown below:
\begin{equation}
\begin{split}
    \widehat{\mathbf{q}_m} &= f_q(\mathbf{e}_m, \tau(m)) \\
    \widehat{\mathbf{k}_n} &= f_k(\mathbf{e}_n, \tau(n)) \\
    \widehat{\mathbf{v}_n} &= f_v(\mathbf{e}_n, \tau(n))
\end{split}
\end{equation}
We impose a condition on $\tau$ that $\forall i > j, \tau(i) > \tau(j)$. This requirement ensures that: (1) No two distinct tokens are assigned the same new position index, and (2) The causality in language modeling remains intact, meaning only query vectors with a larger index can access the key and value vectors with an equal or smaller index.

The concept of position engineering can be also explained through placeholder tokens. Placeholder tokens are defined as tokens that are excluded the computation of attention scores, yet they are allocated position indices. To elaborate, when the calculation of $a_{m,n}$ is undertaken as described in the Equation (\ref{eq:attention_output}), and either the $m$-th or $n$-th token is identified as a placeholder, the conventional computation is bypassed, and $a_{m,n}$ is set to $0$. While placeholder tokens do not directly influence the attention scores at their positions, they do alter the position indices of other input tokens. As depicted in Figure \ref{fig:position engineering}, the insertion of placeholder tokens between sentences 1 and 2 affects the relative positional information between them, which in turn influences the calculation of attention scores between tokens of the two sentences. The connection between the position editing function and the placeholder tokens can be described as follows: Employing a position editing function $\tau$ translates to adding $\tau(i + 1) - \tau(i) - 1$ placeholder tokens after the $i$-th token, and specifically, adding $\tau(0)$ placeholder tokens before the $0$-th token.

\subsection{Position Engineering}

Consider a particular task defined by $(Q, A)$, for which a training set $\{(Q_i, A_i)\}_{i=1}^{N}$ has been sampled according to the task distribution $\Gamma$. We transform each question $Q_i$ into its corresponding text prompt $P_i$. A large language model $\mathcal{M}$ is utilized, which operates based on the prompt $P_i$, and its output is evaluated through a scoring function $r$, denoted as $r(\mathcal{M}, P_i)$. To potentially enhance the performance, a position editing function might be applied to each question prompt. This function is assumed to be parameterized by a vector $\bm{\theta}$, and is denoted as $\tau_{P_i;\bm{\theta}}$. After the application of the positional editing function, a new score is generated, formulated as $r(\mathcal{M}, P_i, \tau_{P_i;\bm{\theta}})$.

For instance, in retrieval-augmented generation (RAG) tasks, the prompt $P_i$ is typically composed of three segments: the instruction, the documents, and the question. It can be possible to define $\bm{\theta}=[\theta_1, \theta_2]$, while $\theta_1$ translates to inserting $\theta_1$ placeholder tokens between the instruction and the document segment, and $\theta_2$ translates to inserting placeholder tokens between the document segment and the question.

Formally, prompt engineering is framed as an optimization problem. We aim at finding the optimal $\bm{\theta}$ that maximizes the score:
\begin{equation}
\begin{split}
   \bm{\theta}^{*} = \mathop{\arg\max}\limits_{\bm{\theta}}{\frac{1}{N}\sum_{i=1}^{N}{r(\mathcal{M}, P_i, \tau_{P_i;\bm{\theta}})}}
\end{split}
\end{equation}
In this research, we utilize a basic algorithm for tackling the optimization problem by initially defining a limited number of candidates for $\bm{\theta}$ and assessing each candidate's score via brute force. Notably, since $\bm{\theta}$ is a numeric vector, the search process can be accelerated by adopting various optimizers, such as Gaussian processes of Bayesian optimization \cite{srinivas2009gaussian}. The exploration of more sophisticated optimization methods will be considered in future works.

\section{Experiments}
\label{sec:experiments}

In this section, we present our experiments and findings for position engineering. We evaluate two prevalent tasks for LLMs, namely Retrieval-Augmented Generation (RAG) and In-context Learning (ICL). Our primary testing model is Llama2-13B-chat \cite{touvron2023llama2}, although we also expand our experiments to include additional models in the Appendix.

\subsection{Position Engineering for RAG}
\label{sec:position-engineering-rag}

\begin{table*}[]
\centering
\begin{tabular}{@{}ccccccc@{}}
\toprule
\textbf{Dataset} & \textbf{N Doc} & \textbf{Baseline} & \textbf{Position Engineering} & \textbf{Abs Impr.} & \textbf{$\bm{\theta_A^{*}}$} & \textbf{$\bm{\theta_B^{*}}$} \\ \midrule
NQ Open          & 1              & 0.341             & \textbf{0.435}                & \textbf{+9.5\%}     & 2,000            & 400              \\
NQ Open          & 3              & 0.424             & \textbf{0.490}                & \textbf{+6.6\%}     & 2,100            & 300              \\
NQ Open          & 5              & 0.452             & \textbf{0.501}                & \textbf{+5.0\%}     & 1,600            & 600              \\ \midrule
EntityQuestions  & 1              & 0.452             & \textbf{0.511}                & \textbf{+5.8\%}     & 1,400            & 500              \\
EntityQuestions  & 3              & 0.501             & \textbf{0.531}                & \textbf{+3.0\%}     & 1,200            & 300              \\
EntityQuestions  & 5              & 0.535             & \textbf{0.558}                & \textbf{+2.3\%}     & 1,300            & 400              \\ \midrule
TrivialQA        & 1              & 0.582             & \textbf{0.657}                & \textbf{+7.5\%}     & 1,300            & 200              \\
TrivialQA        & 3              & 0.646             & \textbf{0.697}                & \textbf{+5.1\%}     & 1,500            & 300              \\
TrivialQA        & 5              & 0.669             & \textbf{0.698}                & \textbf{+2.9\%}     & 2,300            & 200              \\ \midrule
WebQuestions     & 1              & 0.319             & \textbf{0.473}                & \textbf{+15.4\%}    & 1,900            & 500              \\
WebQuestions     & 3              & 0.410             & \textbf{0.507}                & \textbf{+9.7\%}     & 2,100            & 400              \\
WebQuestions     & 5              & 0.434             & \textbf{0.514}                & \textbf{+8.1\%}     & 1,600            & 800              \\ \bottomrule
\end{tabular}
\caption{The test results for RAG. We initially examine all possible combinations to determine the optimal configuration on the training set, which is denoted as ${\theta_A^{*}}$ and ${\theta_B^{*}}$. This optimal configuration is then applied on the test set, and the results are presented in the table. The baseline is $\theta_A=\theta_B=0$. The term "Abs Impr." represents absolute accuracy improvement in percentage. The Llama2-13B-chat model is utilized for the experimentation.}
\label{tab:rag-position-engineering}
\end{table*}


\noindent \textbf{Datasets:} To explore the effectiveness of position engineering on RAG tasks, we utilize four open-domain QA datasets: NQ open \cite{lee-etal-2019-latent}, EntityQuestions \cite{sciavolino2021simple}, TrivialQA \cite{joshi2017triviaqa}, and WebQuestions \cite{berant2013semantic}. These datasets each include a training and an evaluation (or test) set, with each set comprising a series of question-and-answer pairs. From the original training set of each dataset, we randomly select 300 QA pairs to serve as our training set for position engineering. Similarly, we randomly select 2,000 pairs from their original test sets to constitute our test set. In cases where a dataset lack a test set, we utilize its evaluation set instead. The Contriever model, which has been fine-tuned on the MS-MARCO dataset, is employed as the retrieval model \cite{izacard2021unsupervised}. We employ document passages from Wikipedia as our source for retrieval, with each passage containing a total of 100 words \cite{karpukhin2020dense}. $k$ document passages, specifically $k = 1,3,5$, are retrieved, and subsequently concatenated and fed into LLMs. Our evaluation metric is the best exact match accuracy, judging whether any correct answer is in the output, which is a common practice in previous works \cite{kandpal2023large,mallen2022not}. 

\begin{figure}[htbp]
\centering
\includegraphics[width=0.75\linewidth]{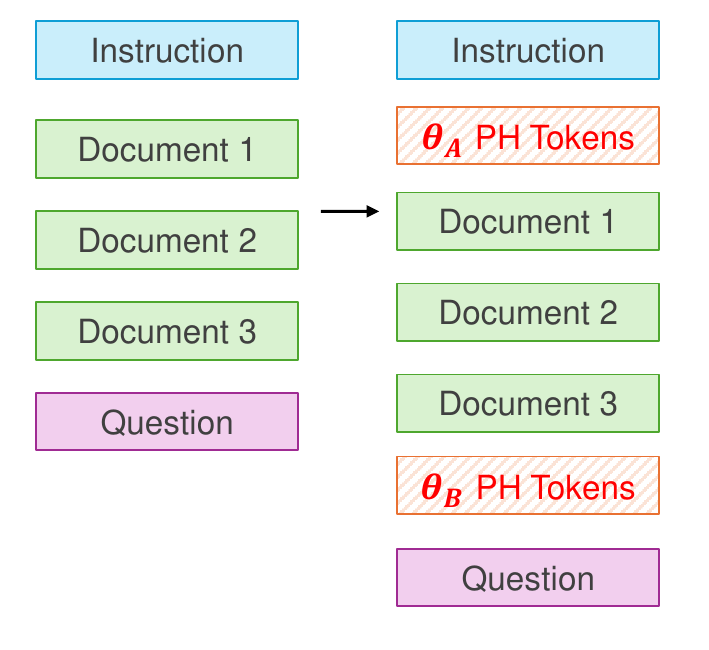}
\caption{Position Engineering for RAG. In the figure, the term "PH tokens" refers to the placeholder tokens introduced in Section \ref{sec:alter-position-information}. We investigate a defined search space, with inserting $\theta_A$ placeholder tokens between the instruction and document segments, and $\theta_B$ placeholder tokens between the document and question segments. Both $\theta_A$ and $\theta_B$ range from $\{0, 100, ..., 2500\}$, subject to $\theta_A + \theta_B \leq 2500$.}
\label{fig:rag-position-engineering}
\end{figure}

\noindent \textbf{Search Space:} We adopt the following prompt template for all RAG experiments. The prompt template is divided into three segments. The first segment provides instructions for the task; the second segment presents a list of retrieved documents, each accompanied by its title and a passage; and the third segment combines the instruction with a specific question. These segments are referred to as the instruction segment, the document segment, and the question segment for convenience.

\begin{tcolorbox}[breakable,title=The prompt template for RAG:]
\label{tb:prompt-rag}
\begin{minted}[fontsize=\small,breaklines,breaksymbolleft=]{text}
Answer the question based on the given documents (some of which might be irrelevant). Only give me the answer and do not output any other words. 

Document (Title: {title}) {passage}

Document (Title: {title}) {passage}

Document (Title: {title}) {passage}

Answer the question based on the given documents (some of which might be irrelevant). Only give me the answer and do not output any other words. 
Question: {question}
Answer:
\end{minted}
\end{tcolorbox}

As presented in Figure \ref{fig:rag-position-engineering}, our study explores the methodology of position engineering for RAG by strategically inserting $\theta_A$ placeholder tokens between the instruction and document segments, and $\theta_B$ placeholder tokens between the document and question segments. To narrow down the search space, the values of $\theta_A$ and $\theta_B$ are limited to a predefined set $\{0, 100, ..., 2500\}$. Additionally, we impose a restriction that $\theta_A + \theta_B \leq 2500$, due to the constraints of the context window size. We evaluate the performance of all combinations on the training set with the Llama2-13B-chat model, and then apply the best configuration to the test set.

\noindent \textbf{Results:} Table \ref{tab:rag-position-engineering} displays the results for RAG, indicating that position engineering substantially enhance the RAG's performance across all settings. The most notable improvement is 15.4\%, observed in the WebQuestions dataset with a single retrieved document. The best-performing parameters, $\theta_A^{*}$ and $\theta_B^{*}$, reveal a consistent trend: $\theta_A^{*}$ tends to be a large number, usually in the range of 1,000 to 2,000, while $\theta_B^{*}$ is a smaller figure, ranging between 200 and 600.

\subsection{Universal Position Configuration for RAG}
\label{sec:universal-rag}

It has been observed that the most effective position configurations, represented as $\theta_A^{*}$ and $\theta_B^{*}$ in Section \ref{sec:position-engineering-rag}, demonstrate a consistent trend across all examined datasets. In this section, we aim to determine a single position setting that can enhance RAG performance universally across different datasets and various numbers of retrieved documents.

Given that absolute accuracy scores vary across datasets, we adopt the percentile value of the accuracy score as a metric to assess each position setting. In this context, we define "experiment setting" as the combination of one dataset and a specific number of retrieved documents, and "position setting" as a specific pair of $\theta_A$ and $\theta_B$. For every experiment setting, we accumulate the scores from all position settings. The effectiveness of each position setting is then evaluated based on its percentile ranking, which varies from 0 to 100, within the experiment setting. Finally, The overall efficacy of a position setting is determined by averaging its percentile rankings across all experiment settings.

\begin{figure}[]
\centering
\includegraphics[width=\linewidth]{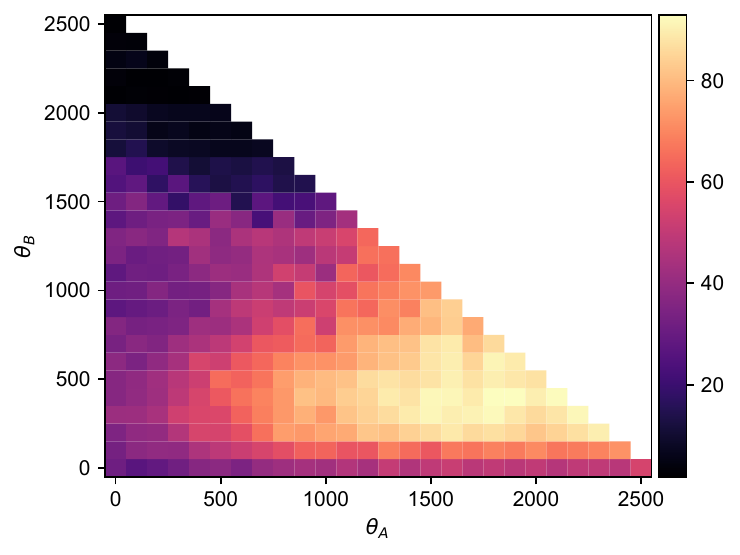}
\caption{We visualize the average percentile values for each positional configuration $(\theta_A, \theta_B)$. These values are initially obtained by aggregating all accuracy scores for a given dataset and a specific number of retrieved documents, and calculate the percentile scores. Subsequently, they are averaged across all configurations, as detailed in Section \ref{sec:universal-rag}.}
\label{fig:rag-analysis}
\end{figure}

The baseline configuration without position engineering ($\theta_A=\theta_B=0$) achieves an average percentile of $31.6$. This suggests that approximately 68\% of configurations can surpass the baseline performance by simply adjusting positional information. The visualization of averaged percentiles for all position settings is provided in Figure \ref{fig:rag-analysis}. 

Generally, it is advantageous to select a $\theta_A$ value within the range of $1300$ to $2000$, and set $\theta_B$ within the range of $300$ to $500$. Setting $\theta_B$ to an excessively high figure (for instance, more than $1500$) significantly deteriorates performance, possibly because it leads to the neglect of document information in prompts. Moreover, for each specified $\theta_B$, an increase in $\theta_A$ is generally associated with better performance.

On the training set, $\theta_A=1900, \theta_B=400$ exhibits the highest percentile value of $92.9$. We apply this configuration to the test set across all datasets and retrieved document numbers. Results presented in Table \ref{tab:rag-universal-gap} demonstrate that it leads to a universal performance improvement. In Appendix \ref{sec:rag-more-models}, we also demonstrate that such configuration remains effective for other models.

\begin{table}[htbp]
\centering
\begin{tabular}{@{}ccc@{}}
\toprule
\textbf{Dataset} & \textbf{N Doc} & \textbf{Abs Impr.} \\ \midrule
NQ Open          & 1              & +9.6\%              \\
NQ Open          & 3              & +7.1\%              \\
NQ Open          & 5              & +4.9\%              \\ \midrule
EntityQuestions  & 1              & +5.6\%              \\
EntityQuestions  & 3              & +3.4\%              \\
EntityQuestions  & 5              & +1.9\%              \\ \midrule
TrivialQA        & 1              & +8.1\%              \\
TrivialQA        & 3              & +4.9\%              \\
TrivialQA        & 5              & +3.2\%              \\ \midrule
WebQuestions     & 1              & +14.8\%             \\
WebQuestions     & 3              & +9.4\%              \\
WebQuestions     & 5              & +9.1\%              \\ \bottomrule
\end{tabular}
\caption{The universal position configuration, $\theta_A=1900, \theta_B=400$, is tested on the test split of all datasets employing the Llama2-13B-chat model. The Table presents the absolute accuracy improvements over the baseline configuration ($\theta_A=\theta_B=0$).}
\label{tab:rag-universal-gap}
\end{table}

\begin{table}[]
\centering
\begin{tabular}{@{}cccc@{}}
\toprule
\textbf{Dataset} & \textbf{N Doc} & \textbf{Baseline} & \textbf{No Inst.} \\ \midrule
NQ Open          & 1              & 0.341             & \textbf{0.353}    \\
NQ Open          & 3              & \textbf{0.424}    & 0.417             \\
NQ Open          & 5              & \textbf{0.452}    & 0.449             \\ \midrule
EntityQuestions  & 1              & 0.452             & \textbf{0.454}    \\
EntityQuestions  & 3              & \textbf{0.501}    & 0.492             \\
EntityQuestions  & 5              & \textbf{0.535}    & 0.532             \\ \midrule
TrivialQA        & 1              & \textbf{0.582}    & \textbf{0.582}    \\
TrivialQA        & 3              & 0.646             & \textbf{0.650}    \\
TrivialQA        & 5              & \textbf{0.669}    & 0.668             \\ \midrule
WebQuestions     & 1              & 0.319             & \textbf{0.335}    \\
WebQuestions     & 3              & \textbf{0.410}    & \textbf{0.410}    \\
WebQuestions     & 5              & 0.434             & \textbf{0.440}    \\ \bottomrule
\end{tabular}
\caption{We test the RAG performance without the instruction segment on the Llama2-13B-chat model. The results are comparable to the baseline, with a slight improvement ranging from 1\% to 2\% on the NQ Open and WebQuestions datasets when a single document is retrieved.}
\label{tab:rag-no-instruction}
\end{table}

\subsection{Without the instruction segment}
\label{sec:ablation-rag}

From Figure \ref{fig:rag-analysis}, it is observed that a larger $\theta_A$ is preferred for optimal performance. $\theta_A$ represents the gap between the instruction segment and the document segment. A larger $\theta_A$ reduces the instruction segment's impact. This raises the question of whether eliminating the instruction segment entirely could further enhance performance. To explore this, we conduct tests, and the outcomes are presented in Table \ref{tab:rag-no-instruction}. It is discovered that the performance of removing the instruction segment is comparable to the baseline setting. The most significant improvement, a 2\% increase, is observed with the WebQuestions dataset when one retrieved document is utilized. However, the enhancement from position engineering in the same experiment setting is 15.4\%. Thus, to achieve the best performance, it is essential to lessen but not eliminate the effect of the instruction segment, a goal that is easy for position engineering, but difficult to accomplish by prompt engineering.


\subsection{Position Engineering for ICL}


\noindent \textbf{Datasets:} To explore the impact of positional engineering on ICL tasks, we employ two datasets: TREC \cite{li-roth-2002-learning,hovy-etal-2001-toward} and SST2 \cite{socher-etal-2013-recursive}. The TREC dataset includes a variety of questions, with the aim being to categorize these questions into 6 coarse and 50 fine-grained question types. We focus on the 6 coarse question types. The SST2 dataset contains movie reviews, with the objective being to categorize these reviews as either positive or negative. For our training set, we randomly choose 300 samples from the original training sets of TREC and SST2. For our test set, we utilize TREC's entire 500-sample test set. For the SST2 dataset, due to the lack of labels in its original test set, we use all 842 samples from its validation set as our test set. For each sample tested, we randomly select 3 examples of each label from the training set as the in-context demonstrations, leading to 18 examples for TREC and 6 for SST2. The exact match score is adopted as the evaluation metric.

\begin{table*}[]
\centering
\begin{tabular}{@{}ccccccc@{}}
\toprule
\textbf{Dataset} & \textbf{Baseline} & \textbf{Position Engineering}  & \textbf{Abs Impr.} & \textbf{$\bm{\theta_A^{*}}$} & \textbf{$\bm{\theta_{mid}^{*}}$} & \textbf{$\bm{\theta_B^{*}}$} \\ \midrule
TREC             & 0.692             & \textbf{0.728} & \textbf{+3.6\%}     & 0                & 40                 & 0                \\
SST2             & 0.915             & \textbf{0.935} & \textbf{+1.9\%}     & 0                & 0                  & 100              \\ \bottomrule
\end{tabular}
\caption{The test results for ICL. We initially examine all possible combinations to determine the optimal configuration on the training set, which is denoted as $({\theta_A^{*}}, \theta_{mid}^{*}, {\theta_B^{*}})$. This optimal configuration is then applied on the test set, and the results are presented in the table. The baseline is $\theta_A=\theta_{mid}=\theta_B=0$. The term "Abs Impr." represents absolute accuracy improvement in percentage. The Llama2-13B-chat model is utilized for this experimentation.}
\label{tab:icl-position-engineering}
\end{table*}

\noindent \textbf{Search Space:} The prompt template provided below is designed for evaluating performance on the SST2 dataset and is divided into three sections: an initial instruction segment that outlines the task, a middle segment that provides examples demonstrating the task, and a final segment that combines the instruction with a query. These segments are referred to as the instruction segment, the example segment, and the query segment, respectively. For the TREC dataset, we employ a similar prompt template, altering only the terms "Review" to "Question" and "Sentiment" to "Question Type" with Llama2-13B-chat.

\begin{figure}[]
\centering
\includegraphics[width=0.9\linewidth]{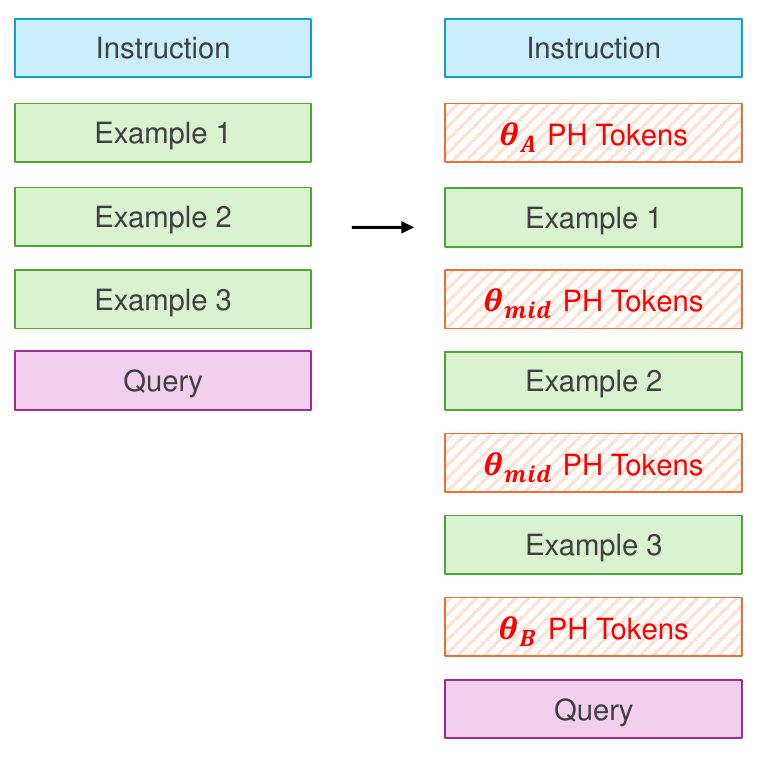}
\caption{Position Engineering for ICL. In the figure, the term "PH tokens" refers to the placeholder tokens introduced in Section \ref{sec:alter-position-information}. We investigate a defined search space, with inserting $\theta_A$ placeholder tokens between the instruction and document segments, $\theta_B$ placeholder tokens between the document and question segments, and $\theta_{mid}$ placeholder tokens among the examples. The candidate value set of $\theta_A$ and $\theta_B$ is set to $\{0, 100, ..., 600\}$, and while $\theta_{mid}$ is set to $\{0, 20, ..., 100\}$.}
\label{fig:icl-position-engineering}
\end{figure}

To investigate the impact of position engineering, we conduct experiments by inserting $\theta_A$ placeholder tokens between the instruction and example segments, $\theta_B$ placeholder tokens between the example segment and the query segment, and $\theta_{mid}$ placeholder tokens among the examples, as depicted in Figure \ref{fig:icl-position-engineering}. The candidate value set of $\theta_A$ and $\theta_B$ is set to $\{0, 100, ..., 600\}$, and while $\theta_{mid}$ is set to $\{0, 20, ..., 100\}$. We evaluate the performance of all possible combinations within the training set and apply the optimal configuration to the test set.

\begin{tcolorbox}[breakable,title=The prompt template for the SST2 dataset:]
\begin{minted}[fontsize=\small,breaklines,breaksymbolleft=]{text}
Please determine the Sentiment of a Review according to the examples below. 

Review: {query}
Sentiment: {label}

Review: {query}
Sentiment: {label}

Review: {query}
Sentiment: {label}

Now, you are given the following Review.
Review: {query}
Please output its Sentiment according to the examples. Only output its Sentiment without outputing anything else.
Sentiment:
\end{minted}
\end{tcolorbox}


\noindent \textbf{Results:} The results for ICL are presented in Table \ref{tab:icl-position-engineering}, indicating an enhancement in performance across both datasets, with an absolute 3.6\% improvement observed on the TREC dataset and an absolute 1.9\% improvement on the SST2 dataset. The optimal position settings, represented as $\theta_A^{*}$, $\theta_B^{*}$, and $\theta_{mid}^{*}$, vary between datasets. Specifically, TREC requires adjusting $\theta_{mid}$ to 40, with $\theta_{A}$ and $\theta_{B}$ set to 0, whereas SST2 requires setting $\theta_{B}$ to 100, with $\theta_{A}$ and $\theta_{mid}$ to 0.

We observe a significant performance drop when $\theta_B$ is set within the $\{200, 300, ..., 600\}$ range, mirroring the trends observed in RAG tasks where a high $\theta_B$ value leads to poor outcomes. $\theta_B$ can be interpreted as a parameter to adjust the impact of the example segment. In the case of SST2, which involves classifying sentiments of reviews—a domain that LLMs might have common knowledge—the choice of $\theta_B^{*}=100$ is intended to slightly reduce the example segment's influence. For TREC, which requires LLMs to learn question types from examples, maintaining $\theta_B^{*}=0$ is optimal.


\section{Discussion}
\label{sec:discussion}

We hypothesize that position engineering serves as a technique to finely adjust the attention weights assigned to different segments within prompts. By extending the positional gap between two segments, the interaction between them is lessened, thereby increasing the attention allocated to other segments. For example, in RAG experiments, an increased value of $\theta_A$ could potentially reduce the impact of the instruction segment while amplifying the attention allocated to the retrieved documents. It is important to note, however, that the initial instruction remains essential, as evidenced in Section \ref{sec:ablation-rag}. Position engineering offers a nuanced approach to adjusting the weights of different blocks without the need for direct addition or removal of text.

Position engineering offers several advantages: (i) It is easier to optimize due to its numerical search space $\{\bm{\theta}\}$, in contrast to prompt engineering, which requires searching over a more complex text space. (ii) It is computationally efficient, as altering position information merely involves updating the position indices input into LLMs, without increasing the overall computational overhead. (iii) It is orthogonal to prompt engineering, meaning the two approaches can be effectively combined.

Future works may advance in the following directions. Firstly, investigating the internal dynamics of LLMs can enhance our understanding of position engineering's underlying mechanisms. Secondly, employing more sophisticated optimizers, such as Gaussian processes or multi-armed bandits, could reduce the search time and discover more refine-grained position editing functions. Finally, the exploration of merging position engineering with prompt engineering could harness the full power of LLMs.

\section{Related Works}
\label{sec:related-works}

\noindent \textbf{Prompt engineering:} Prompt engineering has emerged as a technique to enhance the performance of LLMs by modifying the instructions given to them. For instance, few-shot prompting allows LLMs to learn from demonstrations, a process also known as in-context learning \cite{brown2020language}. Additionally, Chain-of-Thought prompting encourages LLMs to produce intermediate tokens, thereby improving their reasoning capabilities \cite{wei2022chain,kojima2022large}. Another technique, Retrieval-Augmented Generation (RAG), involves retrieving relevant document passages and incorporating them into the prompts \cite{lewis2020retrieval}. It has been discovered that the RAG performance can be improved by adding random documents to the mix of relevant documents \cite{cuconasu2024power}, a technique that is relevant to our study. However, this approach demands significant additional computational resources. In contrast, our proposed method does not require extra computation.

\noindent \textbf{Positional Information in LLMs:} Positional embedding has been introduced to integrate the position information of tokens within the attention layers \cite{vaswani2017attention}. Initially, this concept relied on absolute position indices. However, subsequent developments have introduced methods based on relative positions, such as the relative positional encodings in Transformer-XL \cite{dai2019transformer}, and RoPE \cite{su2024roformer}. ALiBi is a different method for integrating positional information into LLMs \cite{press2021train}, which does not utilize embeddings but introduces a fixed bias based on relative positions during the computation of attention scores. More recent studies have focused on modifying positional embeddings to increase the context window size in LLMs \cite{ding2024longrope,peng2023yarn}. Apart from positional embeddings, the performance of LLMs has been found to correlate with document positions in prompts. In RAG tasks, documents that are positioned in the middle are often more neglected than those at the beginning or the end \cite{liu2024lost}. However, to the best of our knowledge, there has been no similar effort on improving task performance by modifying positional indices.
\section{Conclusion}
\label{sec:conclusion}

In this study, we introduce position engineering, an innovative technique to enhance task performances of LLMs by merely altering the position information in the prompts. Our experimentation with position engineering across a range of tasks and models demonstrates its effectiveness. This approach provides a new avenue for maximizing the capabilities of LLMs.
\section{Limitations}

Our method needs an explicit search process to discover the optimal position setting for a given task. Such search process will cost computation resource and time. Sometimes, the search process can be omitted if a universal good positional setting exists, e.g. the universal setting for RAG tasks with Llama2-13B-chat model. Besides, the internal mechanism of position engineering remains unclear. We hypothesize that position engineering serves as a technique to finely adjust the attention weights assigned to different segments within prompts. Future efforts can be made to further investigate it.

\bibliography{anthology,custom}
\bibliographystyle{acl_natbib}

\appendix
\clearpage
\onecolumn
\section{Appendix}

\subsection{Applying the Universal RAG Configuration to Other Models}
\label{sec:rag-more-models}

\begin{table*}[htbp]
\centering
\begin{tabular}{@{}cccc@{}}
\toprule
\textbf{Dataset} & \textbf{N} & \textbf{Llama2-7B} & \textbf{Mistral-7B} \\ \midrule
NQ Open          & 1          & +8.9\%             & +2.2\%               \\
NQ Open          & 3          & +4.6\%             & +0.2\%               \\
NQ Open          & 5          & +1.0\%             & -0.3\%              \\ \midrule
EntityQuestions  & 1          & +7.1\%             & +0.8\%               \\
EntityQuestions  & 3          & +5.8\%             & -0.1\%              \\
EntityQuestions  & 5          & +0.9\%             & -0.4\%              \\ \midrule
TrivialQA        & 1          & +7.9\%             & +3.1\%               \\
TrivialQA        & 3          & +4.0\%             & +0.7\%               \\
TrivialQA        & 5          & +1.6\%             & +0.5\%               \\ \midrule
WebQuestions     & 1          & +18.5\%            & +3.8\%               \\
WebQuestions     & 3          & +10.0\%            & +2.2\%               \\
WebQuestions     & 5          & +5.9\%             & 0.0\%               \\ \bottomrule
\end{tabular}
\caption{We evaluate the universal position configuration for RAG, as identified in Section \ref{sec:universal-rag} with $\theta_A=1900$ and $\theta_B=400$, across the test splits of all datasets employing the Llama2-7B-chat and Mistral-7B-instruct-v0.2 models. The results showcase the absolute accuracy improvements over the baseline configuration, where $\theta_A$ and $\theta_B$ are both set to 0.}
\label{tab:rag-universal-gap-more-models}
\end{table*}

In Section \ref{sec:universal-rag}, we identified a universal position configuration,$\theta_A = 1900$ and $\theta_B=400$ , on RAG tasks for the Llama2-13B-chat model. In this section, we further investigate whether such configuration remains effective for other models by applying it to the Llama2-7B-chat \cite{touvron2023llama2} and Mistral-7B-instruct-v0.2 \cite{jiang2023mistral} model. The configuration is evaluated on the test splits across all datasets, with the results presented in Table \ref{tab:rag-universal-gap-more-models}. The findings indicate a consistent enhancement in the performance with the Llama2-7B-chat model under the universal position configuration. It is noteworthy that this configuration is initially identified with the Llama2-13B-chat model, suggesting that the Llama2-7B-chat model exhibits similar positional characteristics with Llama2-13B-chat. Furthermore, the Mistral-7B-instruct-v0.2 model also demonstrates consistent performance improvements when utilizing a single retrieved document. However, the performance gains become inconsistent with the use of multiple retrieved documents, indicating a potential need for model-specific adjustments.

\subsection{Applying Position Engineering to Non-RoPE Models}

In our previous evaluation section, Llama2-13B-chat was utilized as the primary model for testing. This model employs RoPE \cite{su2024roformer} to integrate positional information. Furthermore, in this section, we aim to assess the effectiveness of position engineering using models with a different method for incorporating positional information. To this end, we apply position engineering to BLOOMZ-7b1 \cite{muennighoff2022crosslingual} under the same experimental settings for the ICL tasks. BLOOMZ-7b1 is an instruction-fined version of BLOOM \cite{le2023bloom}, which incorporates position information using ALiBi \cite{press2021train}. Unlike RoPE, ALiBi introduces a fixed position-related bias term during the computation of attention scores.

\begin{table*}[htbp]
\centering
\begin{tabular}{ccccccc}
\hline
\textbf{Dataset} & \textbf{Baseline} & \textbf{Position Engineering} & \textbf{Abs Impr.} & \textbf{$\bm{\theta_A^{*}}$} & \textbf{$\bm{\theta_{mid}^{*}}$} & \textbf{$\bm{\theta_B^{*}}$} \\ \hline
TREC             & 0.724             & \textbf{0.782}                & \textbf{+5.8\%}    & 0                & 0                  & 200              \\
SST2             & 0.836             & \textbf{0.946}                & \textbf{+11.0\%}   & 0                & 20                 & 500              \\ \hline
\end{tabular}
\caption{We apply position engineering to the BLOOMZ-7b1 model on ICL tasks. The same search space setting is employed as shown in Figure \ref{fig:icl-position-engineering}. ${\theta_A^{*}}$, $\theta_{mid}^{*}$, and ${\theta_B^{*}}$ is the optimal configuration identified in the training set, which is then applied on the test set. The baseline is $\theta_A=\theta_{mid}=\theta_B=0$. The term "Abs Impr." represents absolute accuracy improvement in percentage compared to the baseline.}
\label{tab:icl-bloomz}
\end{table*}

Specifically, we follow the search space in Figure \ref{fig:icl-position-engineering} for ICL tasks. We determine the optimal position configuration on the training dataset by evaluating all configuration candidates in the search space, subsequently applying this configuration to the test set. Both the training and test sets remain the same with the previous settings. The results are presented in Table \ref{tab:icl-bloomz}. Notably, there is a significant improvement in ICL tasks, with the SST2 dataset showing an absolute improvement of $11.0\%$. It demonstrates that position engineering can be also effective in non-RoPE models.




\end{document}